\documentclass[copyright]{eptcs}
\providecommand{\event}{LoCoCo 2010}
\usepackage{amsmath}
\usepackage{amssymb}
\usepackage{url}
\usepackage{graphics}
\hypersetup{colorlinks=true,linkcolor=black,citecolor=black}

\title{Model Counting in Product Configuration}
\author{Andreas J. K\"ubler \and Christoph Zengler \and Wolfgang K\"uchlin
\institute{Symbolic Computation Group, \\Wilhelm Schickard Institute for 
	Computer Science,\\ Universit\"at T\"ubingen, Germany\\ 
	\url{http://www-sr.informatik.uni-tuebingen.de}}
\email{$\{$kuebler,zengler,kuechlin$\}$@informatik.uni-tuebingen.de}
}
\def\titlerunning{Model Counting in Product Configuration}
\def\authorrunning{A.J. K\"ubler, C. Zengler, W. K\"uchlin}
\begin{document}

\newcommand{\oo}[1]{\operatorname{#1}}
\newcommand{\assign}{\alpha}
\newcommand{\vars}{\operatorname{vars}}
\newcommand{\dom}{\operatorname{dom}}
\newcommand{\MC}{\#SAT}
\newcommand{\PS}{\#P}
\newcommand{\true}{\top}
\newcommand{\false}{\bot}
\newcommand{\impl}{\rightarrow}
\newcommand{\bimpl}{\leftrightarrow}
\newcommand{\df}{d-DNNF}
\newcommand{\ops}{\mathcal{O}}
\newcommand{\vs}{\mathcal{V}}

\maketitle

\begin{abstract}
We describe how to use propositional model counting for a quantitative
analysis of product configuration data. Our approach computes valuable meta
information such as the total number of valid configurations or the relative
frequency of components. This information can be used to assess the severity
of documentation errors or to measure documentation quality. As an application
example we show how we apply these methods to product documentation formulas
of the Mercedes-Benz line of vehicles. In order to process these large
formulas we developed and implemented a new model counter for non-CNF
formulas. Our model counter can process formulas, whose CNF representations 
could not be processed up till now.
\end{abstract}

\section{Introduction}
\label{sec:introduction}

Since R1/XCON~\cite{McDermott:1982} was used by DEC to support computer system
configuration and assembly, product configuration systems have been among the
most prominent and successful applications of AI methods in
practice~\cite{Sabin:1998}. As a result computer aided configuration systems
have been used in managing complex software, hardware or network settings.
Another application area of these configuration systems is the automotive
industry. Here they helped to realize the transition from the mass production
paradigm to present-day mass customization.

Model counting is a technique to count the number of different satisfying
variable assignments of a formula in propositional logic. Up to now model
counting has mostly been used within bayesian
networks~\cite{Sang:2005a,Bacchus:2003} and planning
problems~\cite{Palacios:2005,Kumar:2002}. In contrast, our work aims at using
propositional model counting for a quantitative analysis of comprehensive
product configuration setups.

To demonstrate the applicability of our novel methods, we provide examples
within a well-studied car configuration context. The German automobile
industry follows a build-to-order strategy based on an exceedingly large
product variety. This makes it possible, especially for the manufacturers of
luxury cars, to offer each customer their unique tailor-made car, and it
differentiates the model lines from the mass market. Pil and Holweg
\cite{Pil:2004} have discussed the interconnection of product variety and
order-fulfillment strategies. 

In this context it is interesting for the management to know precisely how
much variety there is in the product line. For the automobile industry, this
question is surprisingly difficult to answer. In particular, sales options
cannot simply be multiplied out because of complex interdependencies
(e.g.~automatic transmission being standard on U.S. exports, and not available
on European cars with small engines). Pil and Holweg published data for the
year 2002, based on an analysis of company material, about customer selectable
variations. In summary, numbers range from the order of $10^{3}$ for the
smallest Peugeot and Nissan models to the order of $10^{8}$ or $10^{9}$ for
the GM Astra and Corsa or the Ford Focus. In the luxury sector, the BMW
3-Series was estimated at $6\cdot 10^{16}$, and the Mercedes C-Class at
$10^{21}$. Finally, for the Mercedes E-Class, 2 body styles, 15 power trains,
285 paint-and-trim options, and 70 other factory-fitted options could be
combined to variations on the order of $3\cdot 10^{24}$. It gets even worse:
Variations induced by sales to countries outside Europe (due to different
emission standards, fuel grades etc.) apparently were not considered by Pil
and Holweg. And customer visible variations are only a subset of the
variations afforded by technical and legal restrictions.

New technology options, such as hybridization, are a continuous source of new
variability, and therefore the issue of rising complexity must have the
ongoing attention of management. Since we work with our industrial partner
Mercedes-Benz on the symbolic verification of configuration
problems~\cite{Kuechlin:2000}, such as whether a given configuration is valid
or not, or which part is required in which configurations, we were able to
investigate \emph{how many valid configurations there are for one specific car
line.} Obviously, there are a number of useful questions associated with this
number. Some of these are: How does the complexity evolve over time? Which
change in complexity would result from adding or canceling an equipment
option? What is the complexity of each body style? Of each engine or power
train configuration? In each country on Earth? For which percentage of the
total variations is a particular part needed? In practice, detailed questions
such as the above can only be answered precisely if the number of variations
can be computed automatically from the manufacturer's configuration data. This
was not possible up to now, which also means that the above numbers were not
verified independently so far.

This paper covers only the first steps towards using model counting in
configuration and manufacturing problems. Clearly, some kind of additional
information about the expected or observed frequency of chosen options in the
real product should be included in the model counts. E.g.~an option can be
allowed in 90\% of all valid configurations, but is chosen only in 10\% of
actual orders. Still, to the best of our knowledge our results represent the
first cases where exact model counts were obtained directly and automatically
from the internal manufacturer configuration data of a current highly complex
product line. We are certain that more ways will be found in which our methods
can be used for computing management information. Also, more extensive data
sets will certainly be produced in the future.

The plan of this paper as follows: In Section~\ref{sec:modelcounting} we
summarize the theory of model counting and give a short overview of current
algorithmic approaches. Section~\ref{sec:applications} presents possible
application fields for model counting in configuration problems. In
Section~\ref{sec:example} we discuss the specific application of our new
methods in the customer order process at Mercedes-Benz. We give a short
description of our new non-CNF model counter \texttt{ncnt} and present
benchmarks of configuration problems used in the current product line.
Section~\ref{sec:futurework} finally summarizes our contributions and points
at some future research directions.

\section{Model Counting}
\label{sec:modelcounting}

Let $\varphi$ be a formula in propositional logic, and let $\vars(\varphi)$
denote the finite set of variables occurring in the formula $\varphi$. An
\emph{assignment} $\assign$ for $\varphi$ is a partial function
$\assign:\vars(\varphi)\to\{\true,\false\}$ mapping variables to truth values
$\top$ and $\bot$. We follow the convention to write $\alpha \models \varphi$
when some formula $\varphi$ holds with respect to $\alpha$.
\emph{Propositional model counting} or {\MC} is the problem of computing the
number of all full assignments $\alpha$ for which $\varphi$ holds,
i.e.~$|\{\alpha \mid \alpha \models \varphi \}|$.

Analogous to SAT, which is the canonical NP-complete problem, {\MC} is the
canonical \PS- complete problem. The complexity class {\PS} is the class of
all problems $p$ for which there exists a non-deterministic polynomial-time
bound Turing machine $M(p)$ such that for each instance $I(p) $ of $p$ there
exist exactly as many computation paths of $M(p)$ as solutions for $I(p)$.
Intuitively {\PS} is the class of counting problems for polynomial-time
decidable problems. According to~\cite{Valiant:1979} even the counting
problems for polynomial-time solvable problems like 2-SAT, Horn-SAT, or
DNF-SAT can be \PS-complete.

In this paper we only deal with \emph{exact model counting} (in contrast to
approximative counting). We distinguish between two different approaches for
exact counting: (1) DPLL-like exhaustive search and (2) knowledge compilation.

The vast majority of successful SAT solvers uses the DPLL
approach~\cite{Davis:1960,Davis:1962}. DPLL is basically a complete search in
the search space of all $2^n$ variable assignments with early cuts in the
search tree when an unsatisfiable branch is detected. DPLL-style model
counters like \texttt{CDP}~\cite{Birnbaum:1999},
\texttt{RelSat}~\cite{Bayardo:2000}, or \texttt{Cachet}~\cite{Sang:2004} are
extensions to existing SAT solvers and require an input formula in CNF. If a
formula $\varphi$ with $n$ variables is not satisfiable, the output is 0. If a
satisfying (and possible partial) assignment $\alpha$ is found, the number of
models for this $\alpha$ is computed with $2^{n - |\alpha|}$ and the algorithm
proceeds to explore the rest of the search tree. There are two important
improvements of this DPLL-based approach. The first one is \emph{component
analysis}~\cite{Bayardo:2000} where one identifies different components $C_1,
\dots ,C_n$ in the constraint graph $G$ of a CNF formula $\varphi$. Let
$\varphi_1,...,\varphi_n$ be the sub-formulas of $\varphi$ corresponding to
the components $C_1, \dots ,C_n$. Then the model count $\MC(\varphi)$ is equal
to $\MC(\varphi_1) \times \dots \times \MC(\varphi_n)$, thus we can calculate
the model count of each component independently. This identification of
components can be performed dynamically while descending into the search tree.
The second improvement is the {\MC} correspondence to clause learning in SAT:
\emph{component caching}~\cite{Sang:2004,Thurley:2006}. Since during the
counting process we often compute counts for the same sub-formulas multiple
times, we cache signatures of sub-formulas and their model count according to
certain caching schemes. Variable selection heuristics as known from SAT have
to be adjusted for {$\MC$} wrt. component analysis and caching: While in SAT
one tries to narrow down the search to one specific solution by intelligently
choosing the branching variables, in {\MC} we try to choose variables where
the according constraint graph is decomposed in various
components~\cite{Sang:2005}.

In the knowledge compilation based approach we convert the formula $\varphi$
into another logical representation such that $\MC(\varphi)$ can be computed
in polynomial time. One well known approach for this is the compilation of
$\varphi$ into a binary decision diagram (BDD)~\cite{Bryant:1986}. Once we
have the BDD, we can count all paths from the root node to the $\true$ labeled
node to get the model count of the formula at hand. Narodytska and Walsh
discussed this approach for configuration problems~\cite{Narodytska:2007}.
However our own experiments with the formulas presented in
Section~\ref{sec:example} showed that for these large formulas, which emerge
in our industrial automotive project, the compilation into BDDs is not
feasible. Another logical representation for propositional formulas, which is
used e.g.~in the system \texttt{c2d}~\cite{Darwiche:2004}, is the
\emph{deterministic decomposable negation normal form}
(\df)~\cite{Darwiche:2001}. A DNNF is an extension of a negation normal form
where for each conjunction $\bigwedge_{i=0}^n f_i$ with sub-formulas $f_0,
\dots, f_n$ it must hold that $\vars(f_i) \cap \vars(f_j)= \emptyset$ for all
$0 \leq i < j \leq n$. In a {\df} we have the additional condition that for
each disjunction $\bigvee_{i=0}^n g_i$ with sub-formulas $G = \{g_0, \dots,
g_n\}$ it must hold that for each $\alpha$ we have $|\{g_i \in G \mid \alpha
\models g_i\}| \leq 1$, i.e.~no sub-formulas share the same satisfying
assignment. Once we have the {\df} $\varphi'$ of a formula $\varphi$ we can
count all models by these two rules: \begin{displaymath} \MC(\bigwedge_{i=0}^n
f_i) = \prod_{i=0} ^n \MC(f_i) \ , \ \MC(\bigvee_{i=0}^n g_i) = \sum_{i=0}^n
\MC(g_i). \end{displaymath} It turns out that a {\df} representation is closer
to the original CNF formula and therefore is easier to compute.

\section{Applications in Product Configuration}
\label{sec:applications}

In this section we will point out possible fields of applications for model
counting in product configuration. We use the definition of a configuration
problem as given in~\cite[Definition 1]{Hadzic:2004}: a configuration problem
is a triple $(\vs,D,\Psi)$ where $\vs$ is a set of variables
$x_1,x_2,\ldots,x_n$, $D$ is a set of their finite domains
$D_1,D_2,\ldots,D_n$ and $\Psi = \{\psi_1,\psi_2,\ldots,\psi_m\}$ is a set of
propositional formulas over atomic propositions $x_i = v$ where $v \in D_i$,
specifying conditions that the variable assignments have to satisfy. For each
formula $\psi \in \Psi$ we have $\vars(\psi) \subseteq \vs$. A \emph{valid
configuration} is an assignment $\alpha$ with $\dom(\alpha) = \vs$ such that
$\alpha \models \bigwedge_{\psi \in \Psi} \psi$.

In this paper we consider the special case where we have only propositional
variables in $\vs$ and hence $D_i = \{\true,\false\}$ for all $1 \leq i \leq
n$. The set $\ops$ is the finite set of all configuration options for a
product. Each variable $x_o \in \vs$ represents a configuration option $o \in
\ops$. The variable $x_o$ is assigned to $\true$ if the option $o$ is chosen,
otherwise it is assigned to $\false$. Following this course, the resulting
formulas $\psi \in \Psi$ are propositional formulas and hence $\varphi =
\bigwedge_{\psi \in \Psi} \psi$ is a propositional formula describing all
valid configurations. We will also refer to $\varphi$ as \emph{product
overview formula (POF)}~\cite{Kuechlin:2000}.

\emph{Remark.} The restriction of the variables $x \in \vs$ to propositional
variables does not limit the expressiveness of our problem description. Since
the domains $D_i$ are finite and we only allow atomic propositions of the form
$x = v$, we can use a reduction~\cite{Bryant:2000} from equality logic to
propositional logic.

\subsection{Number of Valid Configurations}
\label{sub:validconfigurations} 

The first question which naturally arises is the total number of valid product
configurations $v = \MC(\varphi)$. Obviously $2^{|\vs|}$ is an upper bound for
$v$, but in most cases $v \ll 2^{|\vs|}$. Nevertheless this number $v$ can
often demonstrate the sheer complexity of a given product. Subsection
\ref{sub:benzcounting} supports this claim by presenting some of these numbers
for the car lines of Mercedes-Benz.

One can also count valid configurations of a product under certain
preconditions. E.g.~we can force options $P \subset \ops$ to be chosen before
performing $\MC(\varphi)$. This can be achieved by computing $\MC(\varphi
\land \bigwedge_{p \in P} x_p)$. This method can yield important information
about the influence of certain options $o$ on the number of valid
configurations. This information again can be used for special domain-specific
variable heuristics for the SAT solving process of formulas of the application
area at hand. The main idea of state-of-the-art SAT solvers is to narrow down
the search space as fast as possible. Thus when we know a set of variables $X
\subseteq \vs$ representing configuration options $P \subset \ops$ which
reduce the number of models of a given formula to a great extent, we can give
these variables $x \in X$ high activity in the SAT solving process of formulas
representing configuration problems of the same product line.

Preselecting a certain option can also be used to compute the relative
frequency of this option in valid configurations. We compute the frequency
$f(o)$ of a given option $o$ with
\begin{displaymath}
	f(o) = \frac{\MC(\varphi \land x_o)}{\MC(\varphi)}. 
\end{displaymath} 
$f(o)$ can be used as additional information to statistical data about the
frequency of options for demand estimation or process optimization.

\subsection{Rating Errors} 
\label{sub:ratingerrors}

Quite important issues arise when reporting errors. Observations from formal
methods in software verification~\cite{Bessey:2010} tell us that the more bugs
you report, the smaller the probability gets that they will eventually be
fixed. Developers as well as product documentation engineers tend to get
overwhelmed quite quickly by extensive error reports leaving them uncertain
where to start correcting defects.

Model counting might help classifying errors according to their severity. We
consider scenarios in which satisfiability of the input formula $\varphi$
indicates error situations --- hence any satisfying assignment may be
interpreted as a counterexample. In configuration problems such situations
mostly arise when checking mutually exclusive component inclusion. Let $c_{1},
c_{2}\in\vars(\varphi)$ denote binary flags for the inclusion of two mutually
exclusive components. Each assignment satisfying $\varphi\land c_{1}\land
c_{2}$ may be interpreted as a product featuring both mutually exclusive
components at the same time, thus computing $\MC(\varphi\land c_{1}\land
c_{2})$ results in the total number of invalid configurations w.r.t. $c_{1},
c_{2}$. Formulas yielding many such ``counterexamples'' intuitively tend to
fail more likely in practice than formulas yielding only negligible numbers.
However, if there are very few models for a given error, we assume that this
error is more intricate than others.

Measuring error severity using model counters as proposed may however turn out
not to be applicable in every domain. Think of Boolean encodings of
configuration options where each option $o$ is assigned to a value of a finite
integer domain $D$ and a consistency assertion $a(o)$ fixing $o$ to a specific
integer value $i$. If checking the Boolean encoding w.r.t. $a(o)$ yields that
$o$ is not restricted to $i$ but may instead take an arbitrary value of $D$,
model counting will consequently return some large number of invalid
configurations. In case $o$ is some exotic configuration option, using the
computed number as a measure of severity might be misleading.

\subsection{Measuring Documentation Quality}
\label{sub:documentationquality}

Often the individual constraints $\psi \in \Psi$ are reflected in a product
documentation. This product documentation is either automatically produced or
manually maintained by experts. For complex products with thousands of options
this documentation evolves over years. This fact can be observed in the
development of the numbers of valid configurations of different versions of
the documentation for the same product. We made observations where for the
same product the number of valid configurations went down from a magnitude of
$10^{34}$ to a magnitude of $10^{10}$ due to a better and more detailed
documentation (c.f.~Subsection~\ref{sub:benzcounting}).

Model counting can also be used interactively by the documentation engineer.
One can immediately see the impact of adding or changing certain constraints
in terms of valid configurations. E.g.~when adding a new constraint does not
change the number of valid configurations it can be redundant and therefore
can be omitted.

\section{Example: Automotive Product Configuration}
\label{sec:example}

In this section we give examples how to apply propositional model counting to
formulas as used in constructibility checking of customer orders at
Mercedes-Benz. Therefore we first give a brief introduction of the notions
used in the product documentation and of the configuration system used in the
mass production of individually configured, personalized cars.

\subsection{Formula Semantics}
\label{sub:formulasemantics}

For the Mercedes lines of cars, product documentation is done based on
propositional logic. For each vehicle class, there are about 1.500 variables
which represent the configuration options. Processing of customer orders at
Mercedes-Benz is organized as a three-staged process~\cite{Kuechlin:2000}:

\begin{enumerate}
\item The customer's choice of configuration options $P \subset \ops$ with $P
= \{x_{o_1},\ldots,x_{o_{|P|}}\}$ is compiled into an initial assignment
$\alpha = \{ x_{o_{1}} \gets \true, \ldots, x_{o_{|P|}} \gets \true \}$. We
refer to propositional variables representing an option $o \in \ops$ also as
\emph{codes}.

\item The assignment $\alpha$ computed in Step~1 is iteratively extended using
so called supplementary rules. A supplementary rule $S(x) \impl x$ consists of
a conditional part $S(x)$ in propositional logic and a supplementary code $x$.
If $\alpha \models S(x)$ for the current assignment $\alpha$ we extend
${\alpha = \alpha \cup \{x \gets \true \}}$.

\item For each code $x$ there is a constructibility condition in propositional
logic of the form $x \impl C(x)$. The initial assignment $\alpha$ of Step~1,
supplemented in Step~2, is finally checked for constructibility in this third
step. If $\alpha \models \bigwedge_{x \in \dom(\alpha)} C(x)$ holds, the
customer order is constructible otherwise it will be refused.
\end{enumerate}

Let $X$ be the set of all allowed codes for a given line of vehicles, then the
afore mentioned POF is defined as follows

\begin{displaymath}
\oo{POF}(X) = \bigwedge_{x\in X} (S(x) \impl x) \land (x \impl C(x)).
\end{displaymath}
\emph{Example.} Consider as a toy example a vehicle where the configuration
options are three different engines with codes $e_1$, $e_2$, $e_3$, two
different gearboxes $g_1$, $g_2$ and three additional features $a_1$, $a_2$,
$a_3$. Engine $e_1$ must be combined with gearbox $g_1$ ($\oo{sr}_{1}$), $e_2$
must be combined with $g_2$ ($\oo{sr}_{2}$). In a car with $e_3$ and $g_1$
also $a_2$ has to be chosen ($\oo{sr}_{3}$), in a car with $e_3$ and $g_2$, we
must choose $a_3$ ($\oo{sr}_{4}$). The resulting supplementary rules are
\begin{displaymath}
  \oo{sr}_1 = e_1 \impl g_1,\!\!\qquad \oo{sr}_2 = e_2 \impl g_2,\!\!\qquad
  \oo{sr}_3 = e_3 \land g_1 \impl a_2,\!\!\qquad \oo{sr}_4 = e_3 \land g_2 \impl a_3.
\end{displaymath}
There is exactly one engine in a vehicle ($\oo{cc}_1$--$\oo{cc}_4$) and
exactly one gearbox ($\oo{cc}_5$--$\oo{cc}_7$). There is also the condition,
that feature $a_2$ must not be combined with $a_3$ ($\oo{cc}_{8}$). The
resulting constructibility conditions are
\begin{displaymath}
\oo{cc}_1 = \true \impl e_1 \lor e_2 \lor e_3,\!\!\quad \oo{cc}_2 = e_1 \impl \neg e_2 \land \neg e_3,\!\!\quad \oo{cc}_3 = e_2 \impl \neg e_1 \land \neg e_3,\!\!\quad \oo{cc}_4 = e_3 \impl \neg e_1 \land \neg e_2
\end{displaymath}
\begin{displaymath}
\oo{cc}_5 = \true \impl g_1 \lor g_2,\!\!\qquad \oo{cc}_6 = g_1 \impl \neg g_2,\!\!\qquad \oo{cc}_7 = g_2 \impl \neg g_1,\!\!\qquad \oo{cc}_{8} = a_2 \impl \neg a_3
\end{displaymath}
The POF is the conjunction of all constructibility conditions and 
supplementary rules: 
\begin{displaymath}
\oo{POF} = \bigwedge_{i=1}^{8} \oo{cc}_i \land \bigwedge_{j=1}^{4} \oo{sr}_j.
\end{displaymath}

The solutions of the POF represent exactly those customer orders (as a set of
configuration options) which can be built under the configuration constraints.
Hence any model $\alpha$ of the POF (i.e.~$\alpha\models \oo{POF}$) describes
a single valid configuration. Based on the notion of the POF, several
consistency tests such as finding necessary codes, detecting redundant parts
or intersections may be performed quite naturally using SAT
solvers~\cite{Kuechlin:2000}.

Given the variability described above, it is no longer possible to store
separate parts lists (bills of material -- BOM) for each possible order.
Therefore, there is a single BOM for the model line (e.g.~E-Class), where each
part $p$ is associated with a selection condition $P(p)$ in propositional
logic. For each customer order, given as the supplemented assignment $\alpha$,
every part whose selection condition evaluates to true under $\alpha$ is
pulled into the order specific BOM. Hence the set of all parts $p$ occurring
in the BOM is defined as $\{p \mid \alpha \models P(p)\}$.

The BOM is actually structured into so-called positions, each of them grouping
together a set of parts which may be alternatively selected for installation
at the same geometric position of the car (e.g.~the set of available radios).
In addition, there is one fictitious (null-)part which represents the
"`nothing-to-install"' case. Counting the models of the selection formulas for
each part may give an indication of the relative frequency of the parts, and
of the frequency of the null-case, which often represents some kind of
exception.

\subsection{Model Counting in the Mercedes-Benz Scenario}
\label{sub:benzcounting}

In the rest of this paper, we will present some results as described in
Section~\ref{sec:applications} gathered by model counting product
documentation formulas in use at Mercedes-Benz. As those formulas are
originally non-CNF, we developed a new exact propositional model counter that
operates directly upon the raw formulas without prior normalization. It is
based upon a reimplementation of the non-CNF SAT solver introduced by
Thiffault et al.~\cite{Thiffault:2004} and uses the DPLL-style approach to
model counting with connected component detection and caching.

\emph{Remark.} When converting a non-CNF formula $\varphi$ into a CNF formula
$\varphi'$ in order to utilize a CNF model counter, it is important to use a
transformation method which preserves the number of models of $\varphi$,
i.e.~$|\{\alpha \mid \alpha \models \varphi\}| = |\{\alpha \mid \alpha \models
\varphi'\}|$. Consider e.g.~the original formula $\varphi = a \lor (b \land
c)$. To get an equisatisfiable formula $\varphi'$ in CNF we can use the
Tseitin transformation with either implications ($T^\impl(\varphi)$) or
equivalences ($T^\bimpl(\varphi)$):
\begin{align*}
T^\impl(\varphi) ={}& f_{or} \land (f_{or} \impl a \lor f_{and}) \land (f_{and} \impl b \land c) 
={}& f_{or} \land (\neg f_{or} \lor a \lor f_{and}) \land (\neg f_{and} \lor b) \and (\neg f_{and} 
\lor c)\\
T^\bimpl(\varphi) ={}& f_{or} \land (f_{or} \bimpl a \lor f_{and}) \land (f_{and} \bimpl b \land c) 
={}& f_{or} \land (\neg f_{or} \lor a \lor f_{and}) \land (f_{or} \lor \neg a) \land (f_{or} \lor 
\neg f_{and}) \land\\ 
& & (\neg f_{and} \lor b) \land (\neg f_{and} \lor c) \land (f_{and} \lor \neg b \lor \neg c)
\end{align*}
We can now compare the number of models for each formula: $\MC(\varphi) = 5$,
$\MC(T^\impl(\varphi)) = 6$, and $\MC(T^\bimpl(\varphi)) = 5$. We see that
only $T^\bimpl(\varphi)$ preserves the number of models.

\subsubsection{Number of Valid Configurations}

Model counting pure PO formulas without any extensions (i.e.~computing
$\MC(POF)$, the number of different valid orders) demonstrates the vast
multitude of constructible vehicles due to customer choices: We have been able
to compute results of $5.9\times 10^{10}$ up to $9.9\times 10^{103}$
constructible orders for the E-class line of Mercedes-Benz cars (cf. Tab.
\ref{tab1:total}). \texttt{C212\_A}, \texttt{C212\_B}, and \texttt{C212\_C}
indicate different body styles of the same car line. Those impressive numbers
additionally provide an invaluable aid in arguing for formal methods use in
large configuration settings. Our non-CNF model counter has directly been
applied to the non-normalized POFs whereas \texttt{Cachet} and \texttt{c2d}
have been given Tseitin transformed $(T^\bimpl)$ CNF versions as input. Our
comparison states that preserving formula structure by directly reasoning over
the input without prior normalization pays off in scenarios featuring large
input: \texttt{Cachet} timed out after 6 hours on \texttt{C212\_C}
respectively exited abnormally on \texttt{C212\_B}, \texttt{c2d} has been
manually aborted (as it is lacking a timeout option) after 6 hours returning
no results.

\stepcounter{footnote}
\footnotetext{Environment: Linux (2.6.23), 3.4 GHz Intel Pentium D, 2 GB 
memory}
\begin{table}[tdp]
\caption{Computing the total number of orders${}^\thefootnote$}
\begin{center}
\begin{tabular}{lrr@{$\quad$}rr@{$\quad$}rr@{$\quad$}rr}
\hline
& &&\multicolumn{2}{c}{\texttt{ncnt}}&\multicolumn{2}{c}{\texttt{Cachet}}&
\multicolumn{2}{c}{\texttt{c2d}}\\
problem & \#{}vars & \#{}clauses & result & time & result & time & result & 
	time\\
\hline
\texttt{C212\_A} & 4898 & 10289 & $5.9\times 10^{10}$ & $2.57$ & $5.9\times 
	10^{10}$ & $\textbf{0.05}$ & $5.9\times 10^{10}$ & $51.82$\\
\texttt{C212\_B} & 9574 & 40809 & $9.4\times 10^{60}$ & $\textbf{4709.51}$ & 
	bus error & --- & timeout & ---\\
\texttt{C212\_C} & 10637 & 45990 & $9.9\times 10^{103}$ & $\textbf{7915.00}$ 
	& timeout & --- & timeout & ---\\
\hline
\end{tabular}
\end{center}
\label{tab1:total}
\end{table}

In table \ref{tab2:parts} we summarize some computations we performed in order
to analyze the number of valid orders including individual parts of the BOM,
i.e.~$\MC(\oo{POF}\land P(p))$. Each row identifies a position variant (e.g.~a
certain device in the set of available radios), i.e.~a part that might be
installed at one single position in the car. As simultaneously installing
different parts at the same position poses a physical impossibility, position
variants need to be mutually exclusive in order to preserve product
documentation consistency. Using table \ref {tab2:parts} one may compute
inclusion probabilities for each part depending on the corresponding POF. One
can clearly see that for the few variants we computed, variant 20 dominates
the others. These numbers might serve as a basis in applications estimating
future requirements in purchasing parts or for speeding up SAT solving.

\stepcounter{footnote}
\begin{table}[t]
\caption{Computing the number of orders including specific parts using 
\texttt{ncnt}${}^\thefootnote$}
\begin{center}
\begin{tabular}{l@{$\qquad$}rr@{$\qquad$}rr}
\hline
&\multicolumn{2}{c}{\texttt{C212\_B}}&\multicolumn{2}{c}{\texttt{C212\_C}}\\
part $p$ & result & time & result & time\\
\hline
10 & $3.6\times 10^{57}$ & $5160.25$ & $4.1\times 10^{100}$ & $6528.81$\\
20 & $2.7\times 10^{58}$ & $7686.87$ & $2.3\times 10^{101}$ & $7900.62$\\
30 & $5.3\times 10^{56}$ & $4626.25$ & $5.5\times 10^{99}$ & $7620.96$\\
40 & $1.8\times 10^{57}$ & $5530.34$ & $1.9\times 10^{100}$ & $10933.19$\\
60 & $6.0\times 10^{56}$ & $5607.98$ & $6.1\times 10^{99}$ & $8251.84$\\
70 & $6.0\times 10^{55}$ & $1833.41$ & $4.8\times 10^{98}$ & $3840.32$\\
999& $1.2\times 10^{45}$ & $111.87$ & $4.4\times 10^{88}$ & $136.41$\\
\hline
\end{tabular}
\end{center}
\label{tab2:parts}
\end{table}

\footnotetext{Environment: Mac OS X (10.5), 2$\times$4 Core Intel Xeon (2.8 
GHz each), 32 GB memory, using 1 core}

\subsubsection{Rating Errors} 

In practice it is highly desirable to rank the findings of an automated
verifier. In the real world, nothing is perfect, and every corrective action
has a price. If we have an error condition, which is represented by a formula
$\varphi$ being satisfiable, then $\MC(\varphi)$ may give an indication, how
many cars are potentially affected by the error.

As an example consider physically overlapping parts $p_{1}, p_{2}$: If, due to
the $\oo{POF}$, there is a constructible order featuring both, $p_{1}$ and
$p_{2}$, any assignment $\alpha$ with ${\alpha\models \oo{POF} \land P(p_{1})
\land P(p_{2})}$ describes an erroneously constructible order. Thus computing
${\MC(\oo{POF}\land P(p_ {1})\land P(p_{1}))}$ will return the total number of
erroneously constructible orders wrt. $p_{1}$ and $p_ {2}$. Using those
numbers retrieved by model counting one may intuitively classify errors as
follows: Overlaps leading to a high number of constructible orders (thus being
more likely to actually occur in production) are intuitively more severe than
overlaps featuring a negligible number. Experts concerned with fixing
documentary flaws may thus prioritize their work using results produced by
model counting.

\subsubsection{Measuring Documentation Quality}

Exact model counts also give rise to questions about the appropriate product
documentation language and method. We are currently engaged in introducing
symbolic verification methods to detect and help avoid documentation errors.
Knowing the exact model count gives some indication (at least on the level of
management decisions) of the complexity of detecting errors. E.g.~in order to
prove that a part is no longer needed in a 2002 model line, it must be
verified that none of the $10^{24}$ cars needs the part.

Our results furthermore indicate that model counting POFs might serve as a
measure of product documentation maturity (cf.
Subsection~\ref{sub:documentationquality}): Model counting an early
developer's version of \texttt{C212\_A }resulted in $2.9\times 10^{34}$ models
whereas the version in productive use at Mercedes-Benz yields only~${5.9\times
10^{10}}$.

\section{Conclusion \& Future Work}
\label{sec:futurework}

In this paper we introduced new methods for the quantitative analysis of
configuration formulas using model counting. We described methods for gaining
additional information like the total number of valid configurations, the
relative frequency of options in valid configurations, or the severity of
errors. This information can be used to speed up future SAT solving processes
of similar formulas, illustrate the complexity of the problem, or measuring
the documentation quality. Using our non-CNF model counter \texttt{ncnt} we
have been able to compute results for formulas whose CNF representations are
too large for recent model counters to cope with. Our results give proof of
the applicability of our newly introduced analysis methods to industrial-scale
configuration problems.

As our benchmarks show, model counting for large propositional formulas is
still a quite time-consuming job to do. In {\df} compilation static heuristics
based upon hypergraph decomposition have been successfully used to generate
good variable orders for decomposing large and complex formulas
\cite{Huang:2003}. To our knowledge this approach has not yet been used in
non-compilation based settings. Such heuristics should be introduced in
DPLL-style model counting to speed up decomposition.

Moreover, decomposing problems into independent connected components quite
naturally suggests to harness multithreaded CPU architectures by parallelizing
propositional model counting.

\section{Acknowledgement} 

We would like to thank Matthias Sauter (formerly STZ OIT) for providing us
with helpful remarks, advice, and support.

\bibliographystyle{eptcs}
\bibliography{lococo2010}
\end{document}